\documentclass[conference]{IEEEtran}
\IEEEoverridecommandlockouts

\usepackage{cite}
\usepackage{amsmath,amssymb,amsfonts}
\usepackage{algorithmic}
\usepackage{graphicx}
\usepackage{textcomp}
\usepackage{xcolor}

\usepackage{subcaption}
\usepackage{booktabs}
\usepackage{url}
\usepackage{svg}
\usepackage{array}
\newcolumntype{L}[1]{>{\raggedright\arraybackslash}p{#1}}

\newif\ifworkinprogress
\workinprogresstrue 

\ifworkinprogress

\else

\fi
\def\BibTeX{{\rm B\kern-.05em{\sc i\kern-.025em b}\kern-.08em
    T\kern-.1667em\lower.7ex\hbox{E}\kern-.125emX}}
\begin{document}

\title{From Literature to Hypotheses: An AI Co-Scientist System for Biomarker-Guided Drug Combination Hypothesis Generation
}

\author{
\IEEEauthorblockN{
Raneen Younis\textsuperscript{1,2,*},
Suvinava Basak\textsuperscript{1,*},
Lukas Chavez\textsuperscript{3,4,5},
Zahra Ahmadi\textsuperscript{1,2}
}

\IEEEauthorblockA{
\textsuperscript{1}Peter L. Reichertz Institute for Medical
Informatics (PLRI), Hannover Medical School, Hannover, Germany
}

\IEEEauthorblockA{
\textsuperscript{2}Lower Saxony Center for Artificial Intelligence
and Causal Methods in Medicine (CAIMed), Hannover, Germany
}

\IEEEauthorblockA{
\textsuperscript{3}Sanford Burnham Prebys Medical Discovery
Institute, San Diego, CA, USA
}
\IEEEauthorblockA{
\textsuperscript{4}Rady Children’s Institute for Genomic Medicine, Rady Children’s Hospital and Healthcare Center, San Diego, CA, USA
}
\IEEEauthorblockA{
\textsuperscript{5}Department of Medicine, University of California San Diego, La Jolla, CA, USA
}

\IEEEauthorblockA{
\{younis.raneen, basak.suvinava, ahmadi.zahra\}@mh-hannover.de,
lchavez@sbpdiscovery.org
}

\IEEEauthorblockA{
\textsuperscript{*}Equal contribution.
}
}
\maketitle

\begin{abstract}
The rapid growth of biomedical evidence makes it difficult to translate biomarker mechanisms into actionable drug combination hypotheses. We present \textbf{CoDHy}, an interactive AI co-scientist for biomarker-guided hypothesis generation in oncology. CoDHy constructs task-specific knowledge graphs from curated databases and biomedical literature, then combines graph embeddings with agent-based reasoning to generate, validate, and rank evidence-grounded drug combinations. Through a web interface, researchers specify the biomarker, cancer context, and literature scope; inspect supporting evidence and intermediate results; and iteratively refine the generated hypotheses. The demonstration presents CoDHy's end-to-end workflow and shows how researchers can interactively explore and compare mechanistically supported drug combinations while remaining in control of hypothesis prioritization.
\end{abstract}


\section{Introduction}
Biomedical evidence is expanding rapidly across scientific literature and curated databases, yet its scale and fragmentation make it difficult to translate biomarker mechanisms into testable therapeutic strategies. In oncology, researchers face a practical question: \emph{given a biomarker and cancer context, which drug combination should be investigated next?} Answering this question requires connecting heterogeneous evidence, assessing biological plausibility and novelty, and comparing candidate combinations before costly experiments.

Literature-based discovery first showed that implicit biomedical relationships can be uncovered by connecting concepts through intermediate evidence
\cite{swanson1986undiscovered,sybrandt2020agatha}. More recent approaches use semantic representations, knowledge graphs, and large language models (LLMs) to support scientific hypothesis generation
\cite{wang2024scimon,pmlr-v260-shu25a}. Biomedical agents further combine LLM reasoning with external databases and analytical tools
\cite{inoue2025drugagent,gao2025txagent}. However, hypotheses generated directly from unstructured text can be difficult to trace and verify. Existing systems also provide limited support for interactively comparing alternatives, inspecting their evidence, and refining the discovery context.

\begin{figure}[htbp]
    \centerline{\includegraphics[width=0.50\textwidth]{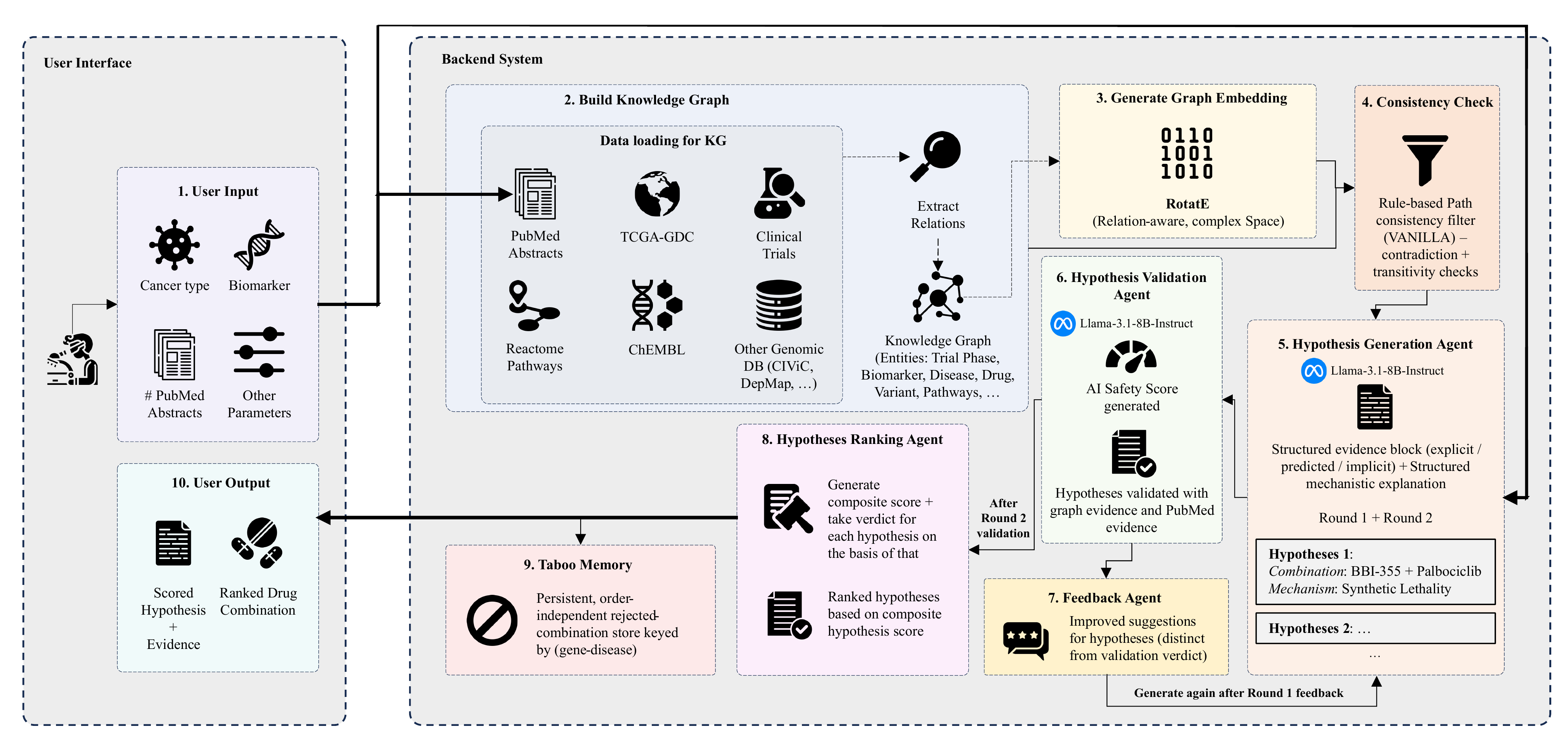}}
    \caption{CoDHy workflow. Users specify a biomarker, cancer context, and literature scope. The system builds a task-specific knowledge graph, generates, validates candidate drug combinations, and returns a ranked list with supporting evidence.}
    \label{fig:system}
\end{figure}

We present \textbf{CoDHy} (\textbf{Co}mbination-\textbf{D}rug
\textbf{Hy}pothesis Explorer), an interactive AI co-scientist for
biomarker-guided drug combination hypothesis generation (Figure~\ref{fig:system}). CoDHy integrates curated biomedical databases with PubMed evidence in a task-specific knowledge graph. It combines graph embeddings and agent-based reasoning to generate, validate, and rank candidate combinations while retaining supporting evidence and mechanistic explanations. Rather than identifying a clinically validated treatment, CoDHy narrows a large hypothesis space to a smaller set of evidence-grounded candidates for researcher assessment and subsequent experimental testing.

Through its web interface\footnote{\url{https://huggingface.co/spaces/suvinavabasak/CoDHy}}, researchers configure the biomarker, cancer type, literature scope, and number of hypotheses. They can inspect and compare ranked candidates, supporting evidence, validation results, and intermediate reasoning, and can revise the input context for reruns. The demonstration presents this end-to-end workflow through oncology use cases, showing how CoDHy helps researchers explore alternative drug combinations and prioritize candidates without replacing expert judgment or experimental validation.
\section{Related Work}
Literature-based discovery identifies implicit biomedical relationships by connecting concepts through intermediate evidence. Systems such as MOLIERE extended this paradigm with semantic and graph-based representations for biomedical hypothesis generation \cite{sybrandt2017moliere}, while recent surveys and benchmarks emphasize structured knowledge, multi-hop reasoning, and explicit validation \cite{yang2025bioverge,kulkarni2025scientific}. Large language models (LLMs) have subsequently enabled literature synthesis and scientific proposal generation \cite{lu2024ai,chen2025ai4research}. Biomedical agents such as DrugAgent and TxAgent combine LLM reasoning with external databases and analytical tools \cite{inoue2025drugagent,gao2025txagent}. Related work also models drug combinations as synergy-prediction tasks; for example, CancerGPT uses an LLM for few-shot drug-pair synergy prediction
\cite{li2024cancergpt}. CoDHy addresses a different task: interactive, biomarker-guided drug combination hypothesis exploration. It combines literature and knowledge-graph evidence with graph-based inference, validation, ranking, and user-controlled discovery settings to narrow the candidate space for expert assessment.
\section{System Design}
\label{sec:system}

CoDHy combines an interactive web interface with the backend pipeline shown in Figure~\ref{fig:system}. Given a biomarker and cancer context, the pipeline builds a task-specific knowledge graph (KG), derives explicit and embedding-based evidence, generates and validates candidate drug combinations. A feedback agent proposes revisions for rejected candidates, which are regenerated and revalidated in second round. The system then returns a ranked list of drug combinations for researcher inspection. CoDHy aims to reduce the hypothesis search space rather than identify clinically validated treatments.

\subsection{Interactive Workflow}


Users specify a biomarker, cancer type, number of PubMed abstracts, language model, and desired number of hypotheses. A public web demo is available online. The interface returns candidate drug combinations with scores, validation verdicts, mechanistic explanations, literature references, and graph-evidence indicators. Users can download the generated report or revise the inputs or evidence scope and rerun, supporting iterative exploration and comparing alternative hypotheses.

\subsection{Evidence Integration and Graph Inference}

For each task, CoDHy integrates curated biomedical databases with literature retrieved from PubMed using the selected biomarker and cancer type. Curated records are represented as typed entities and relations in Neo4j. From the retrieved abstracts, scispaCy~\cite{neumann2019scispacy} extracts biomedical entities, including genes, drugs, pathways, diseases, and outcomes. Extracted relational statements are encoded with \texttt{BAAI/bge-base-en-v1.5} and mapped by cosine similarity to the nearest relation in the KG ontology. This normalization limits uncontrolled relation growth while retaining task-specific literature evidence. Graphs previously constructed for the same context are cached to reduce latency during use.

CoDHy trains \texttt{RotatE} embeddings~\cite{sun2019rotate} over the resulting triples using \texttt{PyKEEN}, a 64D embedding space, \texttt{Adam} optimizor, and 100 epochs. \texttt{RotatE} captures typed relation patterns and supports prediction of plausible unobserved links. The pipeline therefore provides \textbf{three} complementary evidence sources: explicit paths from the KG, inferred links and similar entities from the embedding space. The implementation also supports alternative embedding models for ablation studies.

\subsection{Hypothesis Generation and Refinement}
\label{sec:generation}

The generation agent uses the selected biomarker as an anchor for hybrid graph-retrieval-augmented generation. It retrieves a local subgraph containing explicit biomedical interactions and supplements it with embedding-predicted links and similar entities. These signals are supplied to the language model as separate evidence blocks, with explicit graph evidence given priority over inferred signals.

Before generation, a deterministic consistency filter adapted from VANILLA~\cite{purohit2025vanilla} removes directly contradictory relations and directionally incoherent paths. The generation agent then produces structured hypothesis records containing the drug combination, proposed mechanism, evidence type, supporting references, and a concise explanation linking the retrieved evidence to the candidate combination. This structure supports downstream comparison without treating an LLM's internal reasoning as scientific evidence.

A validation agent assesses each hypothesis for novelty, biological plausibility, feasibility, and potential safety concerns. It searches PubMed for the exact drug combination and uses retrieved reports to support or challenge the proposal. The agent distinguishes direct graph support from embedding-based inference and assigns a verdict of \emph{proceed}, \emph{proceed with caution}, \emph{caution}, or \emph{reject}. These verdicts are prioritization signals for researcher review, not clinical recommendations.

Candidates below a configurable acceptance threshold enter one revision round. A feedback agent identifies weaknesses and supplies targeted revision instructions to the generation agent. A context-specific Taboo List records rejected, order-normalized drug pairs and their rejection reasons, discouraging repeated generation of the same candidates for same inputs. Revised hypotheses are then validated again before ranking.

\subsection{Ranking and Demo Implementation}

For each validated hypothesis, CoDHy computes normalized scores for direct knowledge-graph support, PubMed literature evidence, biological plausibility, safety, toxicity risk, and mechanistic coherence. These are combined using predefined weights into a composite score for ranking. Predefined weights are used because no validated prior exists for weighting the components; learning these weights is left for future work. When available, DrugCombDB~\cite{liu2020drugcombdb} provides reported synergy data. Because CoDHy does not require a cell-line context, it reports the range of synergy scores without interpreting them as evidence. For each candidate, the interface displays its rank, evidence type, references, component scores, and validation verdict.

The Python-based demo is deployed on Hugging Face Spaces. Its backend handles retrieval, KG construction, inference, and agent orchestration, while generation and validation use asynchronous calls to \texttt{Llama-3.1-8B-Instruct} through the Hugging Face Inference API. Results can be inspected in the interface or downloaded as a JSON report, and the source code is publicly available
\footnote{\url{https://github.com/baksho/CoDHy}}.

\section{Evaluation Results and Analysis}
\label{sec:evaluation}

We evaluate CoDHy's ability to narrow the biomarker-guided drug combination search space by generating candidates that are plausible, novel, evidence-linked, mechanistically explained, and testable. We first demonstrate the complete researcher-facing workflow in an oncology use case and then compare CoDHy with five LLMs using a blinded, multi-model judging protocol. The evaluation assesses hypothesis prioritization and presentation; biological efficacy requires subsequent expert and experimental
validation.

\subsection{Setup}
We evaluate seven prespecified biomarker--cancer scenarios, with CoDHy and six \textbf{zero-shot} baselines each generating four drug-combination hypotheses. The baselines are \texttt{Llama-3.1-8B}, \texttt{MedGemma-1.5-4B}, \texttt{Qwen3-8B}, \texttt{JSL-MedLlama-3-8B}, \texttt{OpenBioLLM-8B}, and \texttt{BioMistral-7B}. Because \texttt{Llama-3.1-8B} shares CoDHy's backbone, it serves as a matched zero-shot control to distinguish the contribution of the CoDHy pipeline from that of the underlying model. CoDHy retrieves up to 50 PubMed abstracts per scenario and uses \texttt{Llama-3.1-8B-Instruct} for generation and validation. The complete scenarios, prompts, and structured outputs are available in the repository.
\subsection{End-to-End Use Case}

For \emph{Breast Invasive Carcinoma} with \emph{EGFR} as the focus
biomarker, CoDHy constructs a task-specific knowledge graph and returns
four ranked candidates. Table~\ref{tab:case-study} shows examples to illustrate the evidence types exposed by the interface.

\begin{table}[htbp]
\caption{Example hypotheses returned for the EGFR breast-cancer
scenario. These are research candidates, not treatment recommendations.}
\begin{center}
\renewcommand{\arraystretch}{1.0}
\begin{tabular}{|>{\raggedright\arraybackslash}p{0.18\columnwidth}|
                >{\raggedright\arraybackslash}p{0.28\columnwidth}|
                >{\raggedright\arraybackslash}p{0.14\columnwidth}|
                >{\raggedright\arraybackslash}p{0.18\columnwidth}|}
\hline
\textbf{Combination} &
\textbf{Mechanism} &
\textbf{Evidence} &
\textbf{PMCID(s)} \\
\hline

Afatinib + Fulvestrant &
HER1-2 inhibition + estrogen-receptor modulation &
Explicit KG paths &
PMC6025235 \\
\hline

Osimertinib + Palbociclib &
EGFR resistance reversal + CDK4/6 inhibition &
Embedding-inferred &
PMC7471056 \\
\hline

Lapatinib + Everolimus &
HER1-2/neu receptor inhibition + mTOR inhibition &
Explicit KG paths &
PMC4072025 \\
\hline

Dabrafenib + Trametinib &
BRAF-MEK pathway inhibition &
Embedding-inferred &
PMC12992847, PMC7079252 \\
\hline

\end{tabular}\vspace{-16pt}
\label{tab:case-study}
\end{center}
\end{table}

For each candidate, the interface exposes its proposed mechanism, evidence type, PubMed references, validation verdict, and score. Researchers can compare candidates and modify the inputs for another run. This use case demonstrates the intended role of CoDHy: \emph{narrowing a large search space to a smaller, inspectable candidate set for expert assessment and experimental follow-up.}

\subsection{Blinded LLM-as-Judge Comparison}
CoDHy is compared with six independently prompted zero-shot baselines described in the setup. All receive the same biomarker, cancer context, and output instructions but lack access to CoDHy's knowledge graph, graph embeddings, consistency filter, feedback loop, retrieved evidence, validation pipeline, and ranking procedure.

All outputs are evaluated by the same six-model judge panel: \texttt{MedGemma-1.5-4B}, \texttt{Qwen3-8B}, \texttt{JSL-MedLlama-3-8B}, \texttt{OpenBioLLM-8B}, \texttt{BioMistral-7B}, and \texttt{DeepSeek-R1}, which serves only as a judge. \texttt{Llama-3.1-8B} is a baseline but not a judge. To prevent self-evaluation, each baseline judge abstains from scoring its own outputs. Thus, CoDHy and the Llama baseline are evaluated by all six judges, whereas each other baseline is evaluated by five.

\begin{figure}[htbp]
    \centerline{\includegraphics[width=\columnwidth]{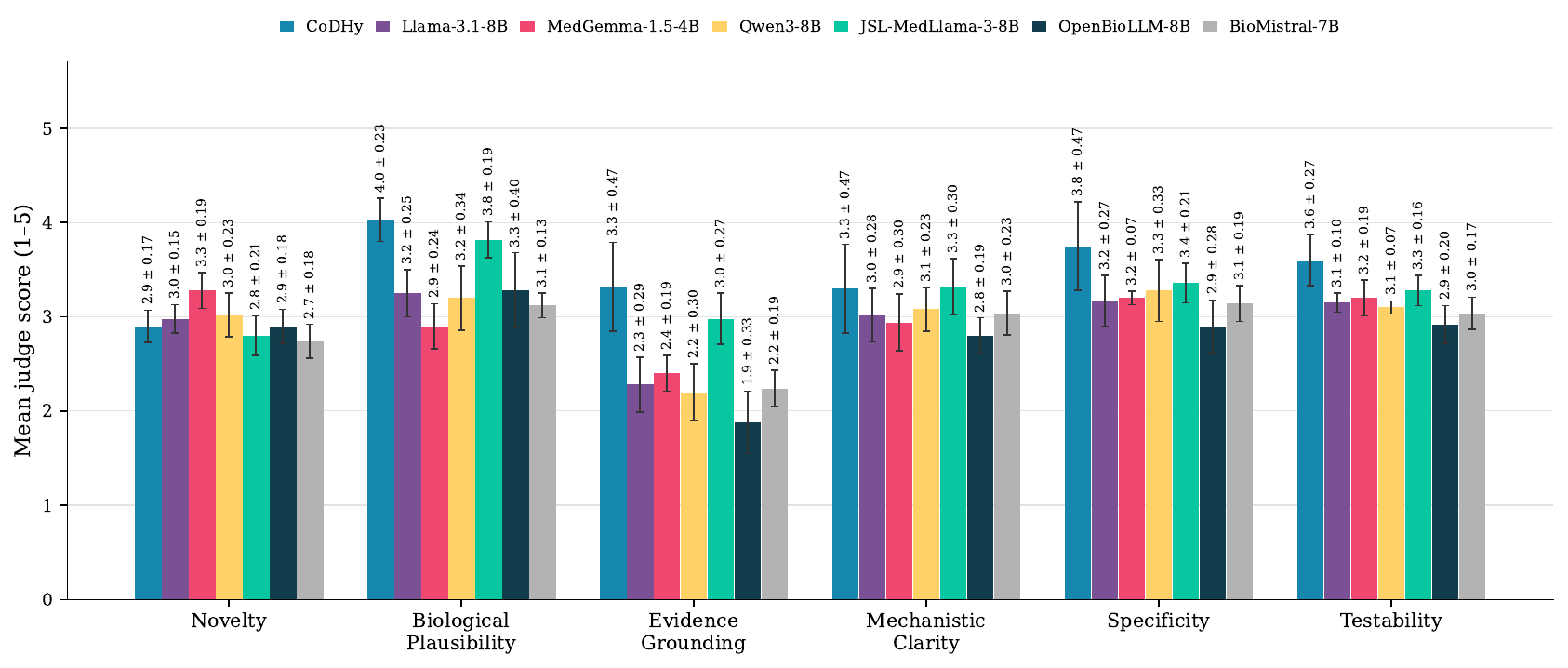}}
    \caption{Mean blinded-judge scores across seven biomarker-cancer scenarios for six evaluation dimensions. \texttt{Llama-3.1-8B-Instruct} is included as a matched zero-shot baseline for CoDHy. Error bars denote standard deviation across eligible judges; baseline generators abstain from judging their own outputs. Higher scores are better.}
    \label{tab:judge-results}
\end{figure}

Judges independently score each eligible hypothesis from 1 to 5 on novelty, biological plausibility, evidence grounding, mechanistic clarity, specificity, and testability. Clinical relevance is excluded because model judgments cannot establish clinical utility. Figure~\ref{tab:judge-results} reports the mean and score spread for each system and criterion across seven scenarios.

CoDHy achieves the highest mean scores for biological plausibility (4.0), evidence grounding (3.3), mechanistic clarity (3.3), specificity (3.8), and testability (3.6). The matched zero-shot \texttt{Llama-3.1-8B} baseline scores 3.0, 3.3, 2.3, 3.0, 3.2, and 3.2 across the six criteria, respectively. CoDHy outperforms it on every criterion except novelty. Because both use the same language model, these gains cannot be attributed solely to model choice and are consistent with a contribution from CoDHy's integrated evidence and reasoning pipeline. Its clearest advantage is evidence grounding, where other baselines score between 1.9 and 3.0. This is consistent with its use of retrieved literature and knowledge-graph evidence, although LLM judges do not independently verify that evidence. For novelty, CoDHy scores 2.9, below \texttt{MedGemma-1.5-4B} (3.3), \texttt{Qwen3-8B} (3.0), and the matched Llama baseline (3.0), but above or equal to the remaining baselines. This may indicate a trade-off between evidence grounding and perceived novelty, but the evaluation does not establish causality.

\begin{figure}[htbp]
\centerline{\includegraphics[width=0.8\columnwidth]{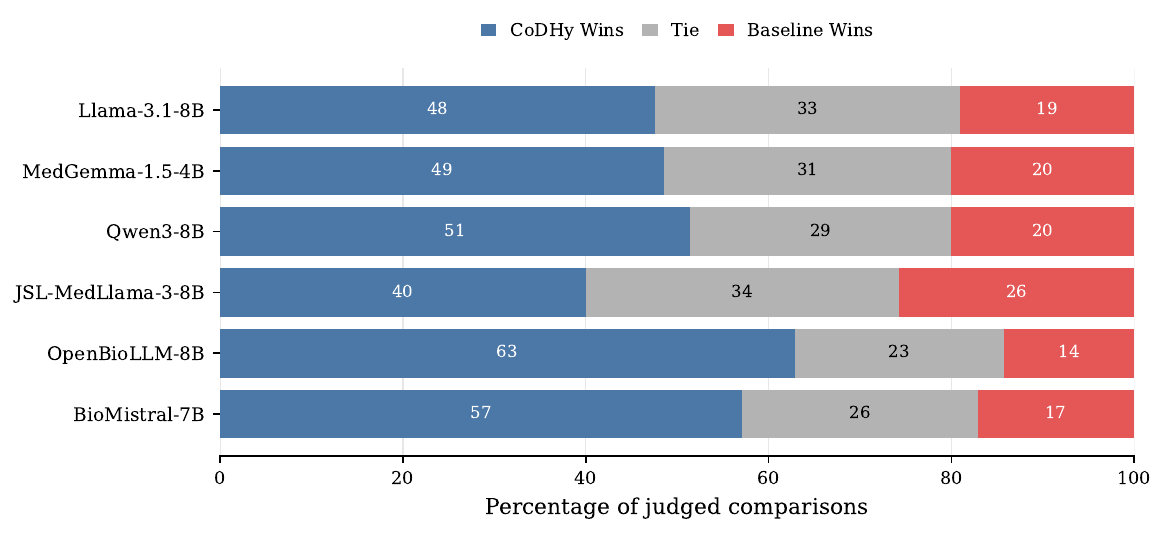}}
    \caption{
    Blinded pairwise preferences for CoDHy versus each zer-shot baseline, aggregated across judges and scenarios. Percentages are reported from CoDHy's perspective.}
    \label{fig:pairwise-results}
\end{figure}

Figure~\ref{fig:pairwise-results} presents a blinded pairwise evaluation across the seven biomarker–cancer scenarios. For each baseline, its hypothesis set and CoDHy's are shown side by side in randomized positions, and judges select the stronger set or a tie. The \texttt{Llama-3.1-8B} comparison is evaluated by all six judges, yielding 42 comparisons. For each of the other five baselines, self-exclusion leaves 35 comparisons. CoDHy is preferred more often in every case, with win rates of 40--63\%, compared with 14--26\% for the baselines.

The repository provides all per-judge scores, pairwise comparisons, agreement statistics, and generated hypotheses. The matched-backbone comparison helps assess the contribution of the CoDHy pipeline. However, LLM judgments reflect perceived hypothesis quality, not biological confirmation; expert review and experimental validation remain necessary.

\section{Conclusions}
We presented CoDHy, an interactive AI co-scientist for biomarker-guided drug combination hypothesis generation. Preliminary results show stronger evidence grounding and favorable pairwise preferences over zero-shot LLM baselines. The matched-backbone comparison with zero-shot \texttt{Llama-3.1-8B} indicates that the observed differences are not solely due to the language model choice and is consistent to the end-to-end CoDHy pipeline. These findings reflect perceived hypothesis quality rather than biological efficacy. Future work will incorporate structured domain-expert validation to improve the reliability and usefulness of the hypotheses.

\section*{Ethical use}
CoDHy uses public biomedical literature and curated databases and
does not process identifiable patient data. Its outputs are research
hypotheses, not clinical recommendations.

\section*{Acknowledgment}
This work was supported by the Federal Ministry of Research, Technology and Space of Germany, project name: EMuLE, grant
number 16IS24059. The last author was partially funded by the CAIMed project, grant number ZN4257.
\bibliographystyle{IEEEtran}
\bibliography{IEEEabrv}
\end{document}